\pdfoutput=1

\documentclass[letterpaper, conference]{IEEEtran}
\IEEEoverridecommandlockouts

\usepackage{times}

\usepackage[sort,numbers]{natbib}
\usepackage{multicol}
\usepackage[bookmarks=true]{hyperref}

\usepackage{amsmath}
\usepackage{amssymb}
\usepackage{cancel}
\usepackage{wasysym}
\usepackage{xfrac}

\usepackage{graphicx}
\usepackage{subcaption}
\usepackage{epstopdf}
\usepackage{float}
\usepackage{tikz}
\usepackage{wrapfig}
\newlength{\sepwid}
\newcommand{\ie}{i.e., }
\newcommand{\eg}{e.g., }
\newcommand{\vs}{vs.\ }

\begin{document}

% paper title
\title{\vspace{6mm}The Soft Landing Problem: Minimizing Energy Loss by a Legged Robot Impacting Yielding Terrain}
% \vspace tip from https://felixduvallet.github.io/blog/ieeetran-robotics/

\author{Daniel J. Lynch, Kevin M. Lynch, and Paul B. Umbanhowar
\thanks{This work was supported by NASA grant NNX15AR24G.

Daniel J. Lynch (corresponding author), Kevin M. Lynch, and Paul B. Umbanhowar are with the Center for Robotics and Biosystems at the Department of Mechanical Engineering, Northwestern University, Evanston, IL 60208 USA. Kevin M. Lynch is also affiliated with the Northwestern Institute on Complex Systems (NICO).
Email: \texttt{daniellynch2021@u.northwestern.edu}, $\left\{\texttt{kmlynch, umbanhowar}\right\}$\texttt{@northwestern.edu}.}
} %end \author

\maketitle

\begin{abstract}
Enabling robots to walk and run on yielding terrain is increasingly vital to endeavors ranging from disaster response to extraterrestrial exploration.
While dynamic legged locomotion on rigid ground is challenging enough, yielding terrain presents additional challenges such as permanent ground deformation which dissipates energy.
In this paper, we examine the \textit{soft landing problem}: given some impact momentum, bring the robot to rest while minimizing foot penetration depth.
To gain insight into properties of penetration depth-minimizing control policies, we formulate a constrained optimal control problem and obtain a bang-bang open-loop force profile.
Motivated by examples from biology and recent advances in legged robotics, we also examine impedance-control solutions to the dimensionless soft landing problem.
Through simulations, we find that optimal impedance reduces penetration depth nearly as much as the open-loop force profile, while remaining robust to model uncertainty.
Through simulations and experiments, we find that the solution space is rich, exhibiting qualitatively different relationships between impact velocity and the optimal impedance for small and large dimensionless impact velocities.
Lastly, we discuss the relevance of this work to minimum-cost-of-transport locomotion for several actuator design choices.
\end{abstract}


\section{Introduction}
Many uses for mobile robots, including disaster response, search and rescue, military ground support, and extraterrestrial exploration,
require locomotion over yielding surfaces, such as soil, sand, snow, gravel, and other regolith.
Given the multitude of legged animals that traverse these yielding substrates with relative ease, legged robots seem a promising alternative to wheeled or treaded robots, which often get stuck in or lose traction on soft ground.

For legged locomotors on yielding terrain, ground deformation due to foot penetration constitutes an irrecoverable energy loss, so minimizing foot penetration depth is relevant to energy-efficient locomotion.
Recent research (\cite{aguilar2016granular,Hubicki,roberts2018reactive,roberts2019mitigating}) has addressed jumping from rest on yielding terrain, \ie the stance-to-flight transition.
Here, conversely, we address the flight-to-stance transition, focusing specifically on the challenge of minimizing foot penetration depth.
We call this the \textit{soft landing problem}.

We consider two control strategies for solving this problem.
First, we formulate the soft landing problem as an optimal control problem to understand properties of control policies that minimize penetration depth.
For robustness to model uncertainty and ease of implementation, we also consider a full-state feedback controller---which renders viscoelastic forces, \ie mechanical impedance, between the robot's body and foot---and we then seek impact-velocity-dependent feedback gains that minimize foot penetration depth.

\subsection{Background}
\subsubsection{Legged locomotors benefit from adjustable impedance}
Animals achieve remarkable metabolic efficiency during running gaits, aided by elastic elements such as tendons and ligaments~\cite{ferris1998running}.
\citet{blickhan1989SLIP} and, later,~\citet{Full} observed that hopping and running gaits could be modeled by a spring-loaded inverted pendulum (SLIP) template, suggesting that elasticity is a defining characteristic of dynamic legged locomotion.
\citet{alexander3springs} proposed that leg springs could reduce the cost of locomotion and that compliant feet could improve ``road holding'' by moderating foot-ground forces during impact.
Similarly,~\citet{ferris1998running} argued that elastic elements with adjustable compliance are crucial to agile locomotion on varied terrain.
Motivated by the need for agility and robustness to terrain uncertainty,~\citet{hurst2004actuator} and, later,~\citet{seok2012actuator} examined the viability of variable-impedance actuators~\cite{english1999antagonistic} for dynamic legged locomotion.

\subsubsection{Soft substrates introduce additional challenges to dynamic legged locomotion}
While there exists considerable and sophisticated research on hard-ground dynamic legged locomotion, research on dynamic legged locomotion on soft ground is in an earlier stage of development.
Nonlinear control synthesis tools such as Hybrid Zero Dynamics (\cite{WesterveltPaper}, \cite{Sreenath}, \cite{HZD_ASLIP}) assume point feet and no slippage in order to model ground contact as a revolute joint between the ground and the foot.
On yielding terrain, this point-contact assumption breaks down.
The stability criterion proposed by~\citet{XiongStability} represents an effort to adapt quasistatic hard-ground locomotion tools (\eg gait generation based on the Zero-Moment Point~\cite{ZMP}) for use on soft ground.
Similarly,~\citet{Hubicki} demonstrated that jumping on granular media could be improved by incorporating a dynamic model of the ground reaction force (GRF) into the robot dynamics used by optimization-based motion planning algorithms.
Although previous studies examined jumping from rest (\eg \cite{aguilar2016granular},~\cite{Hubicki}) and cyclic hopping (\eg\cite{roberts2018reactive},~\cite{roberts2019mitigating}) on yielding substrates, this paper focuses specifically on minimum-penetration-depth landing, which to the best of our knowledge has yet to be addressed.

\subsubsection{Soft substrate ground reaction forces depend on intruder kinematics}
The response of yielding terrains to foot contact varies widely with ground composition, compaction, and inclination, as well as the mass, size, and speed of the locomotor (\eg a sandy beach is a collection of rigid rocks to an ant but is a soft deformable terrain to a human).
Granular media (collections of discrete particles that interact only through repulsion and friction~\cite{GMReview}) are a common material which can be used as a versatile proxy for naturally-occuring soft substrates by tuning their packing density and fluidizing with air~\cite{Qian_2015},~\cite{jin_tang_umbanhowar_hambleton_2019}.
Even with the relative simplicity of a homogeneous granular bed, the resulting GRFs are not trivial, and various models have been proposed to account for their dependence on intruder kinematics (\ie intrusion depth and speed) and particle packing density (see, \eg~\cite{Umbanhowar_cps_2010},~\cite{Li},~\cite{Qian_2015},~\cite{forterre2016}).
For the simplest case of quasi-static vertical intrusion, the GRF increases linearly with penetration depth due to the increase in the frictional force between particles with increasing lithostatic pressure.
However, upon retraction, the GRF drops to nearly zero as the permanently deformed ground does not spring back.

The GRF in granular materials is also velocity dependent, with various models having been proposed depending on the packing density~\cite{Umbanhowar_cps_2010}, the packing density and interstitial fluid~\cite{forterre2016}, and accreted material beneath the foot leading to an ``added-mass'' effect~\cite{aguilar2016granular}.
However, at low impact velocities typical of legged locomotion, the work done by the velocity-dependent GRF term is small relative to the work done by the depth-dependent term.
Given the dominance of the depth-dependent force at these velocities and the number of terrain-specific parameters required to model velocity dependent forces, in this paper we will focus on first-order depth-dependent GRFs in the interest of model generalizability and tractability.

\subsection{Paper Outline}
The remainder of this paper is structured as follows: Section~\ref{sec:model} derives a dimensionless dynamic model for a vertically-constrained two-mass robot impacting soft ground.
We first approach the soft landing problem through analytical optimal control methods in Section~\ref{sec:PMP}.
This approach provides insight into the soft landing problem but yields a brittle optimizer, so in Section~\ref{sec:impedance_optimization} we examine impedance control as a more robust alternative, using simulation to study how the optimal impedance varies with dimensionless impact velocity and model parameters.
In Section~\ref{sec:bang_bang_vs_imp}, we compare the optimal impedance control and bang-bang force control solutions.
We present experimental results for impedance control in Section~\ref{sec:experiments} and discuss extensions of this work to minimum cost-of-transport hopping in Section~\ref{sec:discussion}.

\section{Modeling}\label{sec:model}
\subsection{Soft Ground Model}
\citet{Li} show that resistive force theory gives rise to a depth-dependent GRF in the case of vertical quasistatic penetration of a flat-bottomed intruder into granular media.
While higher-order effects (inertial drag~\cite{KatsuragiH.2007Uflf} and granular accretion~\cite{aguilar2016granular}) are present, we focus on the dominant first-order depth-dependent stress in the interest of generalizability, and note that this approximation is in good agreement with our experimental results (see Section~\ref{sec:experiments}).
Consequently, we model the GRF for soft ground, $f_g$, as 
\begin{equation}\label{eq:one-way_spring_grf}
f_g = 
\begin{cases}
0 & \text{if }q_f \geq 0 \text{ or } \dot{q}_f > 0\textrm{ \textbf{(flight)},}\\
-k_g q_f &\text{if }q_f < 0 \text{ and }\dot{q}_f < 0\textrm{ \textbf{(yielding)},}\\
\left[0, -k_g q_f\right] & \textrm{otherwise \textbf{(static)},}
\end{cases}
\end{equation}
where $k_g > 0$ is the ground stiffness, $q_f$ is the foot position, and $\dot{q}_f$ is the intruder velocity.
This unidirectional ground-spring model is depicted in Figure~\ref{fig:two_mass_force_control_spring_ground}.
The first case (``flight'') simply says the GRF is zero when the intruder is not in contact or is breaking contact.
We refer to the second and third cases as the ``yielding'' and ``static'' regimes, respectively.

\subsection{Robot Model}
\begin{figure}[t]
\centering
\smallskip
\includegraphics[width=0.6\linewidth]{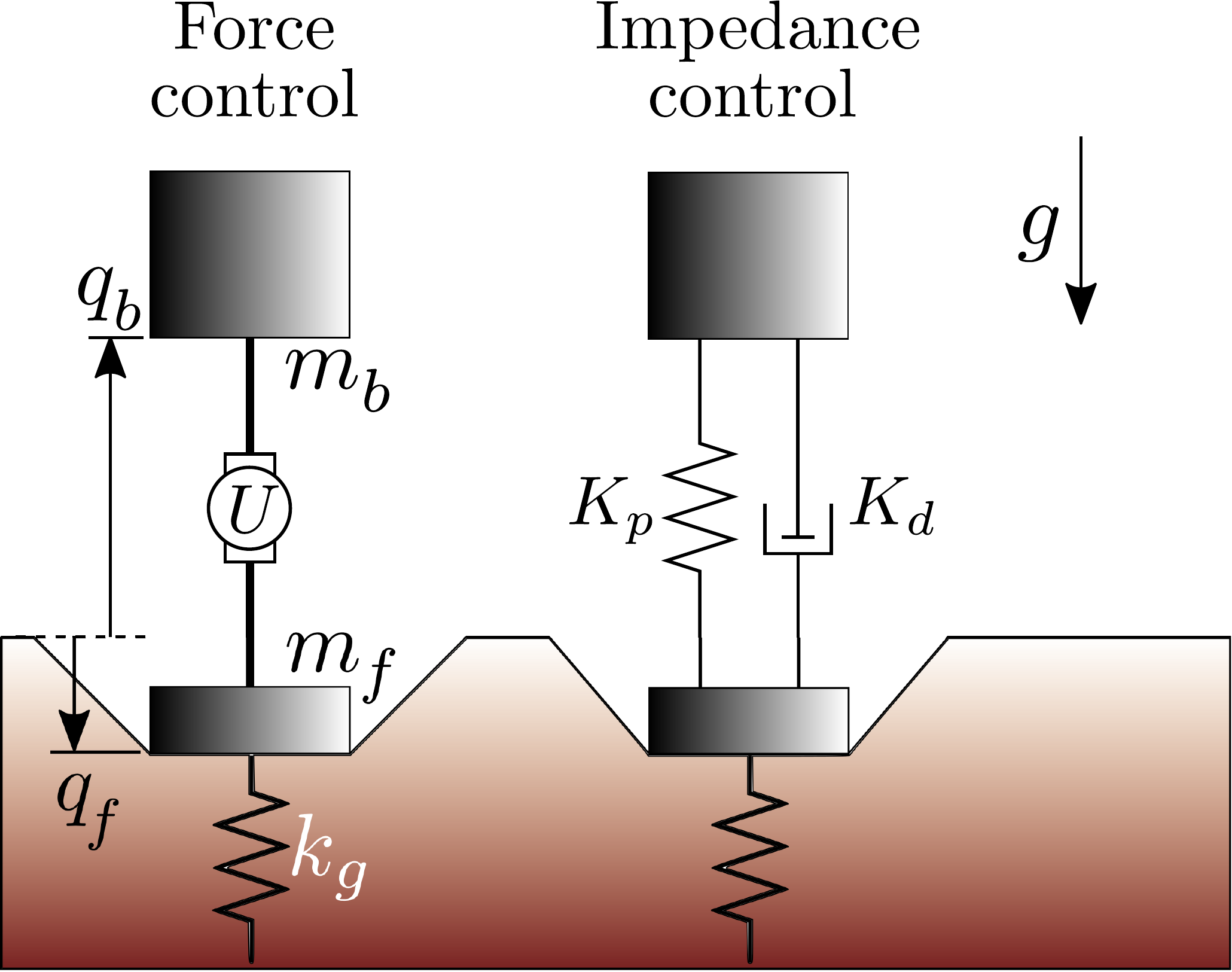}
\caption{Force control and impedance control models (Equations~\eqref{eq:model_ode_2nd_order}-\eqref{eq:dim_imp}).
Soft ground is treated as a unidirectional spring with stiffness $k_g$ (Equation~\eqref{eq:one-way_spring_grf}).
The body, located at $q_b$, has mass $m_b$, and the foot, located at $q_f$, has mass $m_f$.
Heights $q_b$ and $q_f$ are measured relative to the undisturbed ground surface.
}
\label{fig:two_mass_force_control_spring_ground}
\end{figure}

In order to focus on foot-ground interaction, our robot is intentionally simple.
As shown in Figure~\ref{fig:two_mass_force_control_spring_ground}, the robot consists of a body (position $q_b$, mass $m_b$) and a flat-bottomed foot (position $q_f$, mass $m_f$), with a linear motor (idealized as a force source $U$) located between the body and the foot, such that $U > 0$ pushes the two masses apart.
In the case of impedance control, also shown in Figure~\ref{fig:two_mass_force_control_spring_ground}, this force results from stiffness $K_p$ and damping $K_d$ between the body and the foot.

\begin{figure}[t]
\centering
\smallskip
\includegraphics[width=0.6\linewidth]{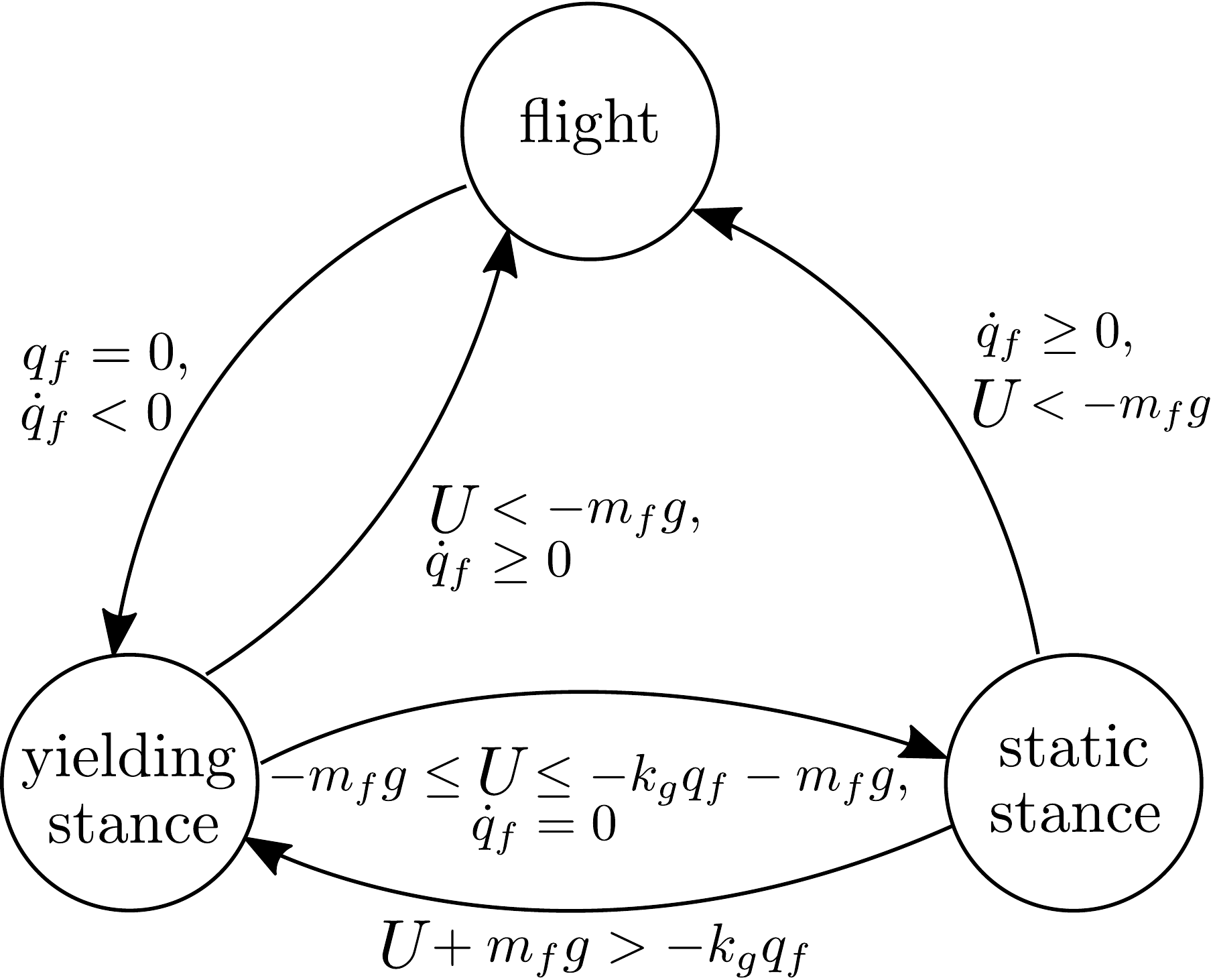}
\caption{Finite state machine depicting transitions between flight, yielding stance, and static stance for the robot and unidirectional spring GRF model given by Equations~\eqref{eq:one-way_spring_grf} and~\eqref{eq:model_ode_2nd_order}.}
\label{fig:FSM}
\end{figure}

The robot dynamics are divided into three phases: flight, yielding stance, and static stance.
We restrict our analysis of the soft landing problem to the two stance phases, beginning at impact ($t = 0, q_f = 0$).
The stance dynamics are
\begin{subequations}\label{eq:model_ode_2nd_order}
\begin{align}
\frac{\mathrm{d}^2 q_b}{\mathrm{d}t^2} &= -g + \frac{U}{m_b}\text{,}\label{eq:body_ode_2nd_order}\\
\frac{\mathrm{d}^2 q_f}{\mathrm{d}t^2} &= -g +\frac{f_g - U}{m_f}\text{,}\label{eq:foot_ode_2nd_order}
\end{align}
\end{subequations}
where the state-dependent GRF $f_g$ is given by Equation~\eqref{eq:one-way_spring_grf}.
We choose $g = 9.81\ \textrm{m}/\textrm{s}^2$ without loss of generality.
Figure~\ref{fig:FSM} shows the finite state machine that describes transitions between yielding stance, static stance, and flight.

For realism, we limit the actuator stroke,
\begin{subequations}\label{eq:stroke_def}
\begin{equation}
D \leq q_b - q_f \leq D + S\text{,}
\end{equation}
where $D$ is some minimum separation distance between the body and the foot.
We choose $D = 0$ without loss of generality.
We also limit the actuator force,
\begin{equation}
-U_{\textrm{max}} \leq U \leq U_{\textrm{max}}\text{.}
\end{equation}
\end{subequations}
In impedance control, viscoelastic forces are rendered through the feedback law
\begin{equation}\label{eq:dim_imp}
U = -K_p\left(q_b - q_f - L_0\right) - K_d\left(\dot{q}_b - \dot{q}_f\right),
\end{equation}
where $K_p > 0$ and $K_d > 0$ represent stiffness and damping, respectively, and $L_0$ is the rest length of the virtual spring emulated through the feedback law above.
We take $L_0 = S/2$.
Throughout this paper, we assume the following initial conditions:
\begin{equation}
q_b(0) = L_0\text{, }
q_f(0) = 0\text{, and } 
\dot{q}_b(0) = \dot{q}_f(0) = V_0,
\end{equation}
where $V_0$ is the impact velocity.

\subsection{Nondimensionalization}\label{sec:nondimensionalization}
To reduce the dimensionality of the model parameter space, we nondimensionalize the robot-ground model in Equations~\eqref{eq:model_ode_2nd_order},~\eqref{eq:stroke_def}, and~\eqref{eq:dim_imp}.
We first introduce the following dimensionless variables for the body and foot positions, time, and the control force: $x_b = q_b/x_s$, $x_f = q_f/x_s$, $\tau = t/\tau_s$, and $u = U/u_s$.
The corresponding unit distance, time, and force are $x_s = m_t g/k_g$ (where $m_t = m_b + m_f$), $\tau_s = \sqrt{m_t/k_g}$, and $u_s = m_t g$.\footnote{Note that there are other valid choices for the dimensionless variables; all that is required is that they span the fundamental dimensions of the system (in this case, mass, length, and time).
Our choice of dimensionless variables is particularly convenient for examining penetration depth because the position scaling factor $x_s$ is the minimum depth at which the ground can support the total weight of the robot.
Moreover, leg stiffness (in the impedance control case) is now measured in units of ground stiffness, so the dimensionless dynamics are independent of the ground stiffness.}
Substituting these expressions for position, time, and control into Equations~\eqref{eq:body_ode_2nd_order} and~\eqref{eq:foot_ode_2nd_order} yields the dimensionless body and foot dynamics
\begin{subequations}\label{eq:nondim_odes}
\begin{align}
\frac{\mathrm{d}^2 x_b}{\mathrm{d}\tau^2} &= \ddot{x}_b = -1 + \frac{1 + r_m}{r_m}u\text{, and}\label{eq:nondim_body_ode}\\
\frac{\mathrm{d}^2 x_f}{\mathrm{d}\tau^2} &= \ddot{x}_f = -1 -\left(1 + r_m\right)\left(x_f + u\right)\text{,}\label{eq:nondim_foot_ode}
\end{align}
\end{subequations}
where the mass ratio $r_m = m_b/m_f$ is the ratio of the body mass to the foot mass.
By defining the state vector $x = \left[x_1,x_2,x_3,x_4\right]^{\top} = \left[x_b, \dot{x}_b, x_f, \dot{x}_f\right]^{\top}$, the nondimensionalized dynamics can be represented in control-affine first-order form $\dot{x} = f(x) + g(x)u$:
\begin{equation}\label{eq:nondim_state-space_dynamics}
\frac{\textrm{d}}{\textrm{d}\tau}\begin{bmatrix}
x_1\\
x_2\\
x_3\\
x_4
\end{bmatrix} =
\underbrace{\begin{bmatrix}
x_2\\
-1\\
x_4\\
-\left(1 + r_m\right) x_3 - 1
\end{bmatrix}}_{f(x)} + 
\underbrace{\begin{bmatrix}
0\\
\frac{1 + r_m}{r_m}\\
0\\
-\left(1 + r_m\right)\\
\end{bmatrix}}_{g(x)}
u\text{.}
\end{equation}
The dimensionless actuator stroke and force limits are
\begin{subequations}\label{eq:nondim_stroke_def}
\begin{equation}\label{eq:nondim_stroke_lim}
0 \leq x_b - x_f \leq s\text{,}
\end{equation}
where $s = S/x_s$, and
\begin{equation}\label{eq:nondim_force_lim}
-u_{\textrm{max}} \leq u \leq u_{\textrm{max}}\text{,}
\end{equation}
\end{subequations}
where $u_{\textrm{max}} = U_{\textrm{max}}/u_s$.
The nondimensionalized impedance-rendering feedback controller is
\begin{equation}\label{eq:nondim_imp_control}
u(x) = -k_p(x_b - x_f - \ell_0) - k_d(\dot{x}_b - \dot{x}_f),
\end{equation}
where $k_p = K_p/k_g$ and $k_d = K_d/\sqrt{m_t k_g}$ are dimensionless stiffness and damping constants and $\ell_0$ = $s/2$.
The dimensionless initial conditions are
\begin{equation}\label{eq:nondim_ICs}
x(0) = \left[\ell_0,v_0,0,v_0\right]^{\top}\text{,}
\end{equation}
where $v_0 = V_0\tau_s/x_s$.

\section{Optimal control formulation}\label{sec:PMP}
We formulate the soft landing problem as a constrained optimal control problem:
\begin{itemize}
\item cost function: $J(x,u) = -x_f(T)$, where $T$ is the free terminal time at which the foot stops intruding.
\item dynamic constraints: $\dot{x} = f(x) + g(x)u$, as defined in Equation~\eqref{eq:nondim_state-space_dynamics}.
\item state inequality constraints: actuator stroke limits, as defined in Equation~\eqref{eq:nondim_stroke_lim}, represented by the vector inequality $h_1(x) \leq 0_{2\times1}$.
\item control bounds: actuator force limits, as defined in Equation~\eqref{eq:nondim_force_lim}, represented by the scalar inequality $h_2(u) \leq 0$.
\end{itemize}
Additionally, there are two constraints on the terminal state $x(T)$.
The first terminal constraint requires the foot to stop, so $\dot{x}_f(T) = 0$.
The penetration depth $x_f(T)$ must support the constant force $u_b$ required to bring the body to rest, given the body velocity $\dot{x}_b(T)$ and remaining stroke $x_b(T) - x_f(T)$:
\begin{equation}
u_b = \frac{r_m}{1 + r_m}\left(1 + \frac{\dot{x}_b^2(T)}{2\left(x_b(T) - x_f(T)\right)}\right).
\end{equation}
The resulting second terminal constraint is
\begin{equation}\label{eq:terminal_state_equality_constraint_2}
x_f(T) + \frac{1}{1 + r_m} + u_b = 0.
\end{equation}
We use Pontryagin's Maximum Principle (PMP,~\cite{PMP}) to determine the structure of the penetration-minimizing force profile $u^*(\tau)$.
We first define a control Hamiltonian:
\begin{equation}
H = \lambda^{\top}\left(f(x) + g(x)u\right)\text{,}
\end{equation}
where $\lambda(\tau) \in \mathbb{R}^{4}$ is the state of the adjoint system, propagating backwards in time from $T$.
In the presence of bounded controls, PMP states that the optimal control $u^*(t)$ satisfies
\begin{equation}
u^*(\tau) = \underset{-u_{\textrm{max}} \leq u \leq u_{\textrm{max}}}{\text{arg max}} H(x^*(\tau),u(\tau),\lambda(\tau)).
\end{equation}
The state inequality constraints and terminal constraints increase the complexity of the problem and prevent us from obtaining an analytical expression for $u^*(\tau)$, but PMP allows us to make several key observations about the control Hamiltonian $H$:
\begin{itemize}
\item The dynamics are time invariant, so $H$ is constant.
\item The terminal time $T$ is free, so $H(T) = 0$, and because $H$ is constant, then $H(\tau) = 0\ \forall \tau \in \left[0,T\right]$.
\item The control Hamiltonian $H$ is linear with respect to the control $u$.
\end{itemize}
In light of the facts above, Hamiltonian maximization implies bang-bang control---the control force $u$ is always at one of its boundaries---as long as neither state inequality constraint is active for a finite period of time~\cite{MaurerH.1977OOCP}.
These conditions---for which bang-bang control minimizes penetration depth---remain true for any robot-ground dynamic model that is time-invariant and control-affine, including the inertial drag~\cite{KatsuragiH.2007Uflf} and added-mass~\cite{aguilar2016granular} GRF models.

\begin{figure}[t]
    \centering
    \smallskip
    \includegraphics[width=0.7\linewidth]{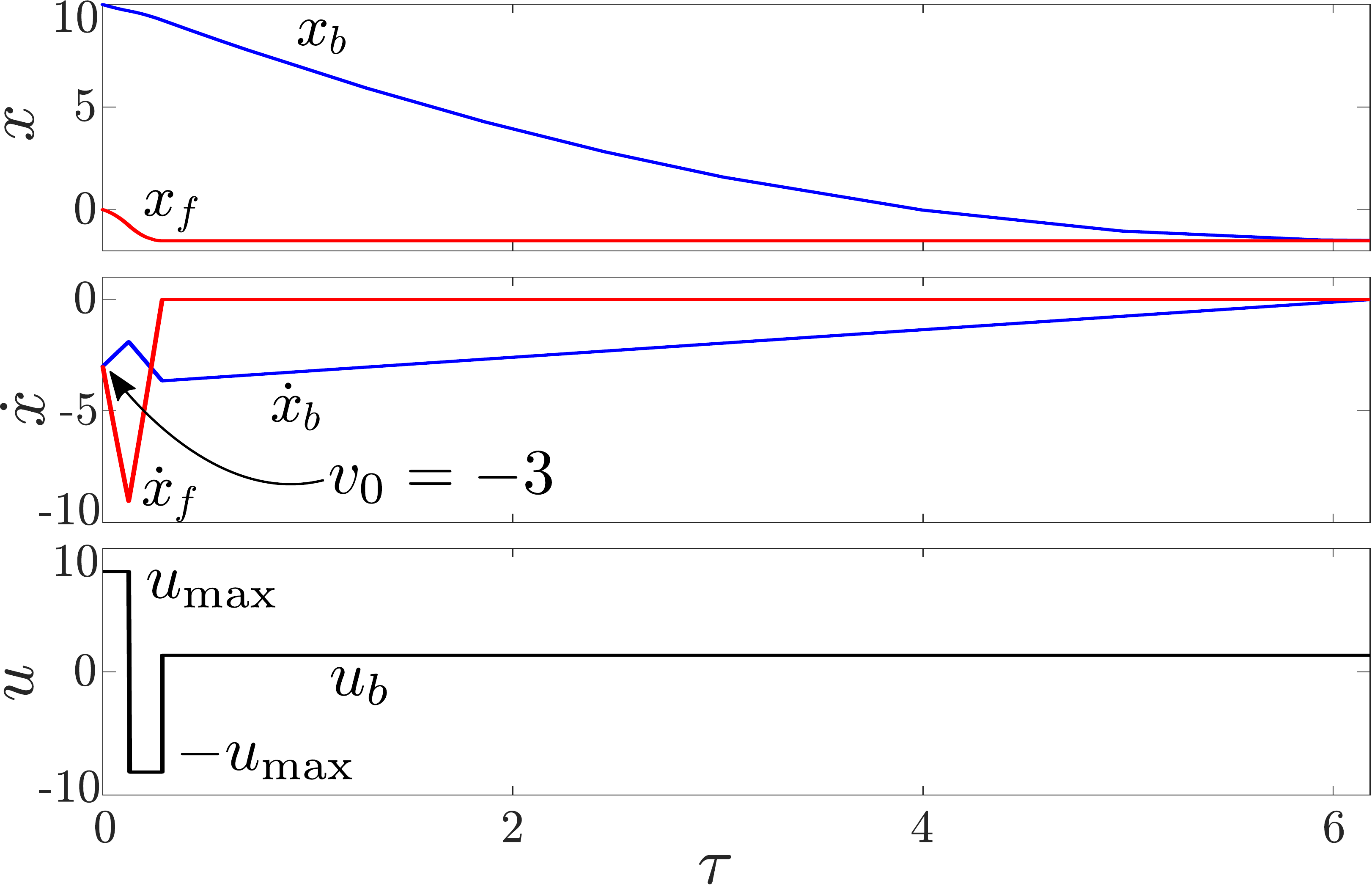}
    \caption{Example robot state trajectory under bang-bang control, $v_0 = -3$, $r_m = 5$, $s = 20$, $u_\textrm{max} = 8.2$.
Bang-bang control quickly ($\tau\approx0.5$ in this example) stomps the foot to the depth required to support the body-arresting force $u_b$.}
    \label{fig:ex_bb_traj}
\end{figure}

Figure~\ref{fig:ex_bb_traj} shows a typical optimal bang-bang force profile and the resulting motion if we assume that there is a single control switch, from $u_\textrm{max}$ to $-u_\textrm{max}$, before the foot comes to rest.
We solve numerically for the optimal switch time $\tau^*$.
Intuitively, this force profile first stomps the foot down into the ground then pulls up on it to stop its descent.
In this way, the robot as quickly as possible deforms the ground to the depth that will support the force needed to bring the body to rest.
By performing this ``stomp'' quickly, the robot has more time and therefore more stroke to decelerate the body, and therefore does not need to penetrate as deep.\footnote{Note that penetration depth can be further minimized by increasing the pre-impact leg extension $x_b(0) - x_f(0)$, subject to stroke limits, in order to increase the distance over which the body must decelerate and thereby reduce the required penetration depth.
Such initial condition optimization is exhibited in nature, \eg by falling cats as they prepare for impact~\cite{falling_cats}.}
We also considered additional switching events for force control (using MATLAB's \texttt{fmincon} solver to find the optimal switching times) and found the difference in penetration depth to be no larger than $1\%$;
while a single switching event may not always be optimal, it appears quite close to optimal.

Bang-bang solutions to the soft landing problem appear to reduce penetration depth by at least a factor of two, compared to a rigid impactor (see Figure~\ref{fig:imp_vs_bb_vs_rigid} and the discussion in Section~\ref{sec:bang_bang_vs_imp}).
While open-loop bang-bang control appears to minimize foot penetration depth, the absence of feedback and the discontinuities in applied force $u$ result in a brittle optimizer that is difficult to implement on real hardware.
For robustness and ease of implementation, we consider an impedance control solution to the soft landing problem in the following two sections.

\section{Numerical impedance optimization}\label{sec:impedance_optimization}
In this section, we examine the effect of dimensionless leg stiffness $k_p$ and damping $k_d$ on penetration depth $x_f$ and seek optimal pairs $\left(k_p^*, k_d^*\right)$ that minimize penetration depth, given the stroke $s$ and mass ratio $r_m$, for a range of impact velocities $v_0$.
We hypothesize that the optimal stiffness and damping will be less than one, or in other words, that the optimal impedance will not be stiffer than the ground.
We use MATLAB's \texttt{ode15s} integrator to numerically simulate impedance-controlled impacts for $0 \leq k_p \leq 1$, $0 \leq k_d \leq 1$, and $-10 \leq v_0 \leq 0$, and initially take $r_m = 5$ and $s = 20$.
These values approximately describe our experimental apparatus, detailed in Section~\ref{sec:experiments}.

\begin{figure}[t]
\smallskip
\centering
    \begin{subfigure}[b]{0.43\linewidth}
        \centering
        \includegraphics[width=\textwidth]{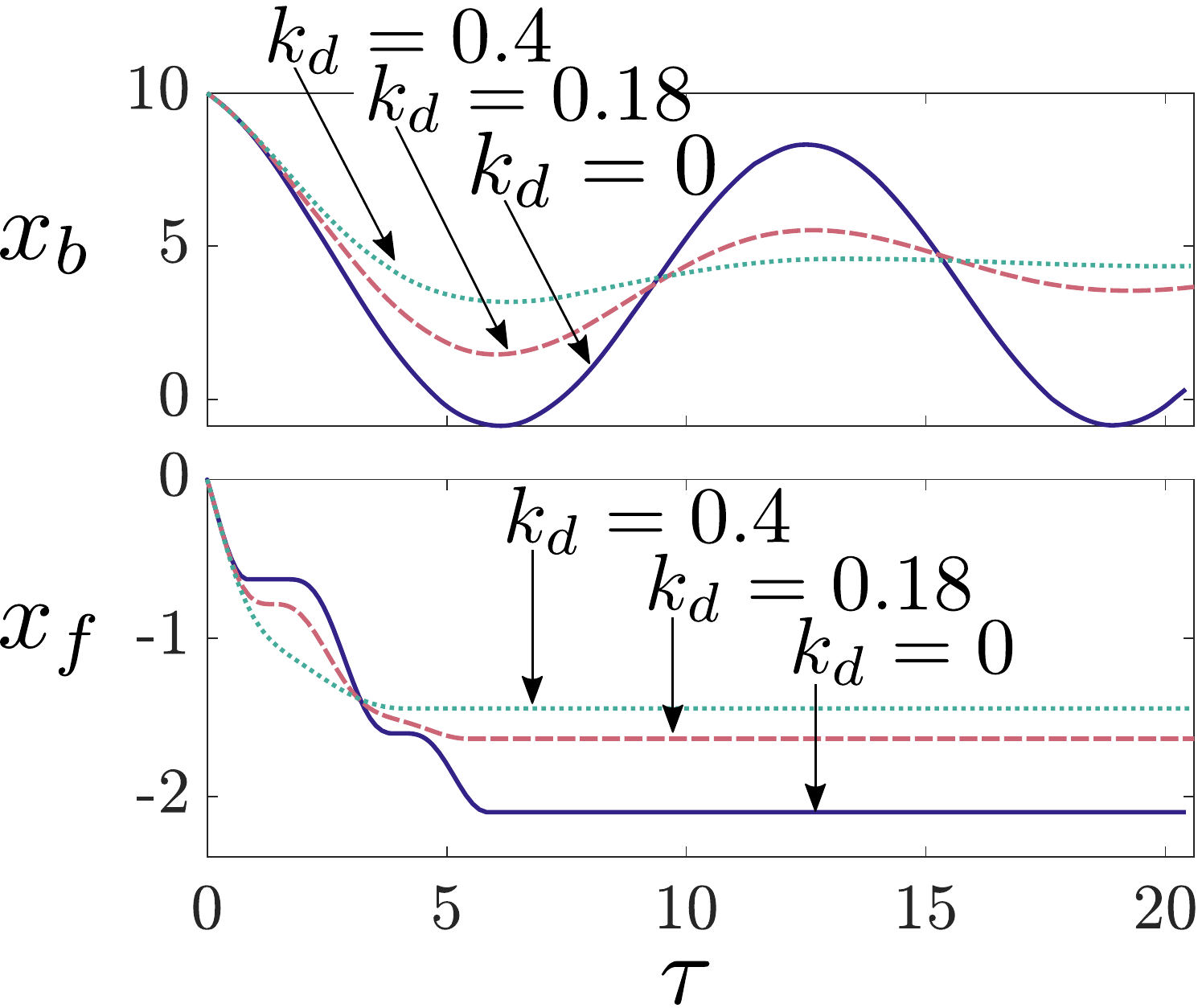}
        \caption{Stepped-intrusion trajectories.}
        \label{fig:multi_intrusion_trajectories}
    \end{subfigure}%
    ~
    \begin{subfigure}[b]{0.52\linewidth}
        \centering
        \includegraphics[width=\textwidth]{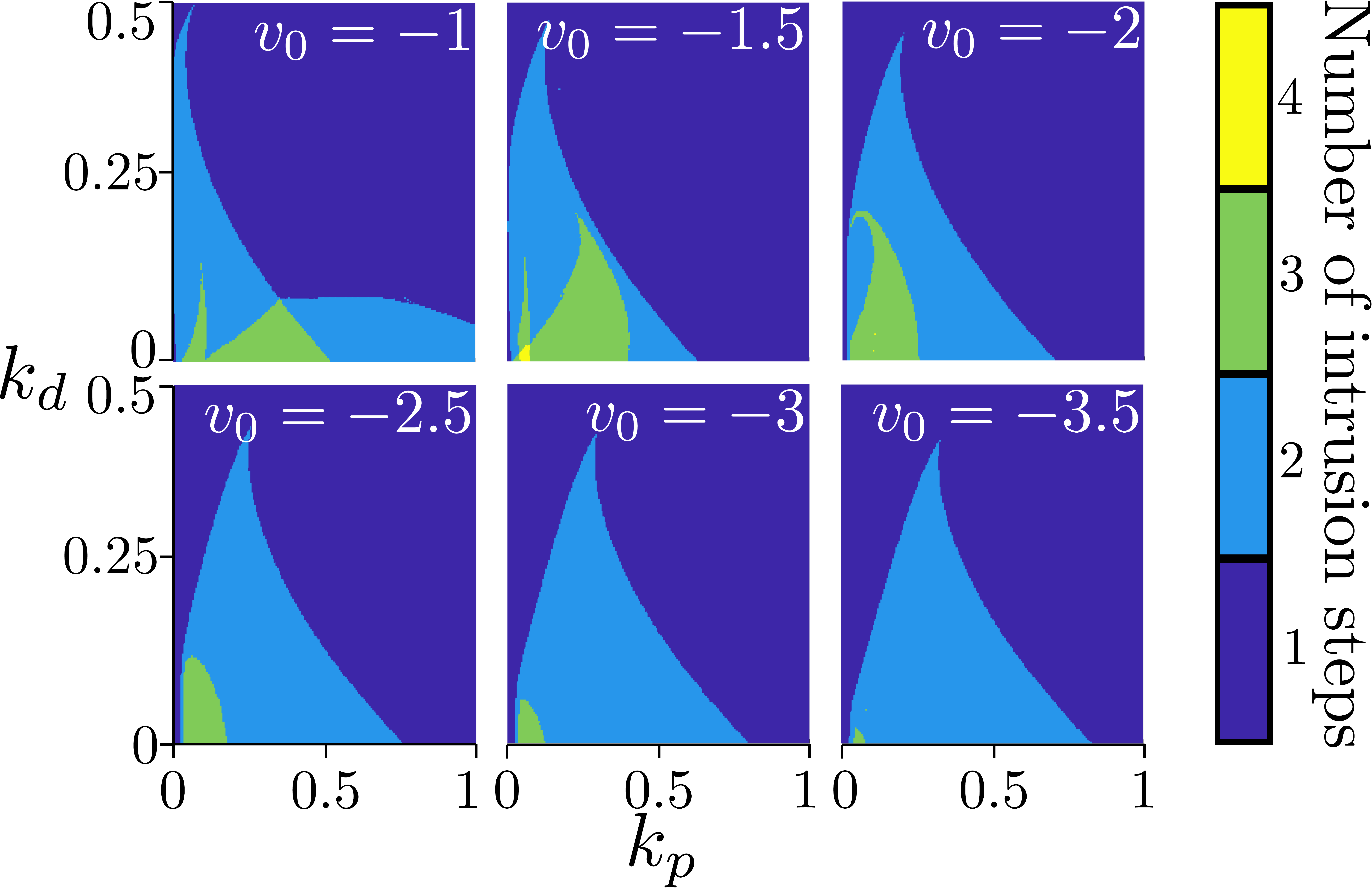}
        \caption{Stepped-intrusion regions in the $k_p$-$k_d$ plane for several $v_0$.}
        \label{fig:multi_intrusion_grid}
    \end{subfigure}
\caption{\textbf{a) } For impact velocity $v_0 = -1$ and stiffness $k_p = 0.2$, three different values of damping result in one-step intrusion ($k_d = 0.4$), two-step intrusion ($k_d = 0.18$), and three-step intrusion ($k_d = 0$).
\textbf{b) }Number of intrusion steps \vs dimensionless stiffness, damping, and impact velocity; $r_m=5$, $s=20$.
Multi-step intrusions appear to occur primarily at low stiffness and low damping.}
\label{fig:multi_intrusion_regions}
\end{figure}

Figure~\ref{fig:multi_intrusion_trajectories} shows three example trajectories, obtained by holding $k_p$ constant while varying $k_d$.
For certain values of $k_p$ and $k_d$, the robot foot stops multiple times before reaching the final depth.
The foot initially comes to rest at some depth then, due to the impedance $\left(k_p, k_d\right)$ and the body motion $\left(x_b,\dot{x}_b\right)$, the downward force on the foot exceeds the ground yield threshold at that depth, and the foot resumes intrusion.
Figure~\ref{fig:multi_intrusion_grid} shows several snapshots of these stepped-intrusion regions emerging, morphing, and vanishing as $v_0$ grows.

\begin{figure}[t]
\centering
\includegraphics[width=0.75\linewidth]{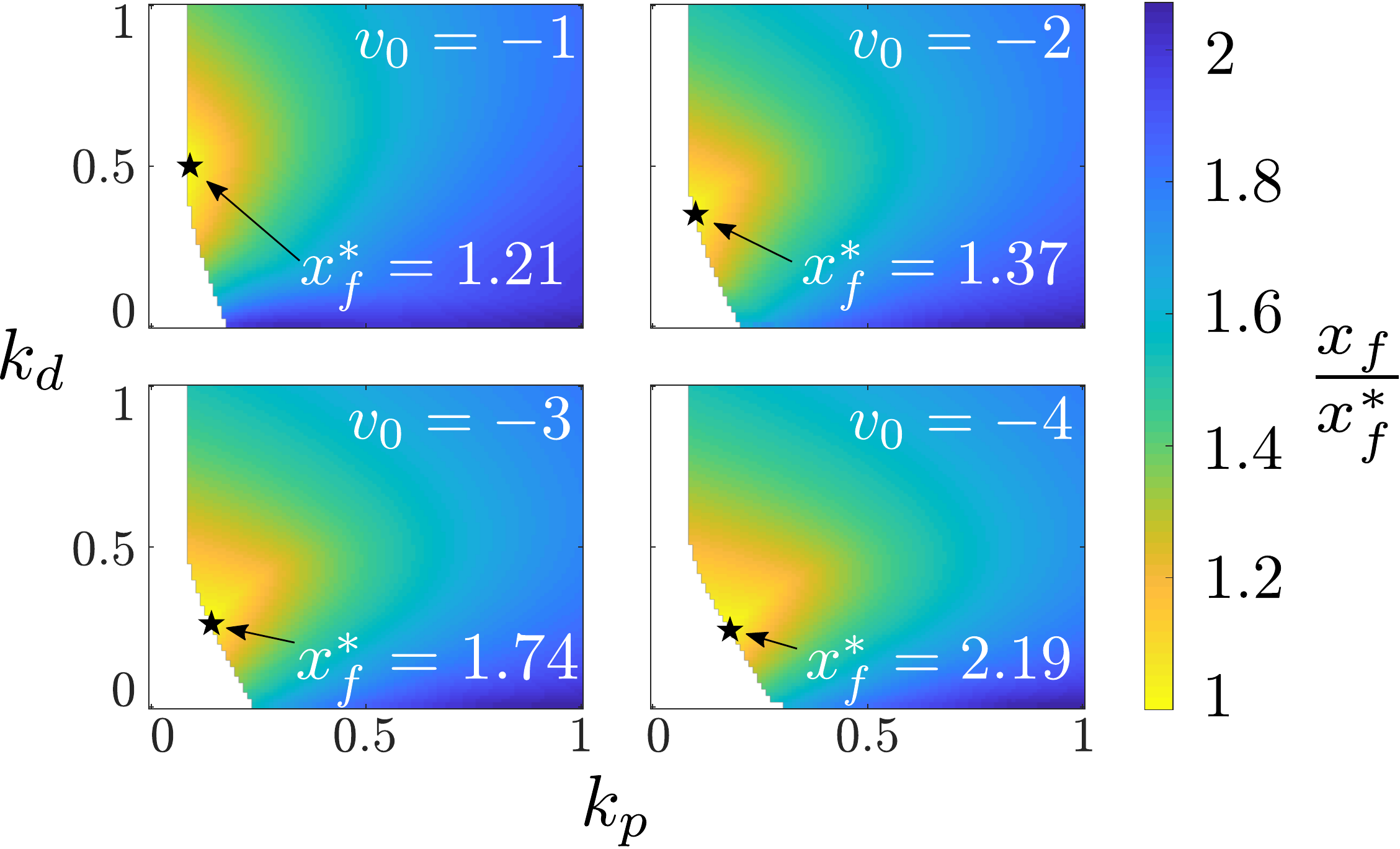}
\caption{Normalized dimensionless penetration depth $x_f/x_f^*$ \vs stiffness $k_p$ and damping $k_d$ for several impact velocities $v_0$ ($r_m = 5, s = 20$).
Minimum penetration depth is denoted by $x_f^*$.
The white region on the left corresponds to impacts where the stroke limit is exceeded.
As $v_0$ grows in magnitude, $\partial x_f/\partial k_p$ (sensitivity of penetration depth to relative stiffness) also grows in magnitude.
}
\label{fig:Kp_Kd_grid}
\end{figure}

Allowing for arbitrarily many steps during intrusion, we seek the depth-minimizing impedance for a given impact velocity, mass ratio, and stroke limit.
Figure~\ref{fig:Kp_Kd_grid} shows snapshots of the impedance-depth relationship for several impact velocities.
Impacts where the stroke limit $0 \leq x_b - x_f \leq s$ was violated are discarded from the search for $\left(k_p^*,k_d^*\right)$.
Note that in many instances, impedances that result in stepped intrusions also violate the stroke limit, but in some cases the optimal impedance results in repeated intrusions.
For all simulated $v_0$, the optimizer (denoted by a $\star$ in Figure~\ref{fig:Kp_Kd_grid}) resides on the stroke limit boundary.
While this increases the brittleness of the optimizer, a safety factor can be simply added by reducing the stroke limit used in control computations compared to the actual stroke limit.

\begin{figure}[t]
    \centering
    \smallskip
    \includegraphics[width=0.95\linewidth]{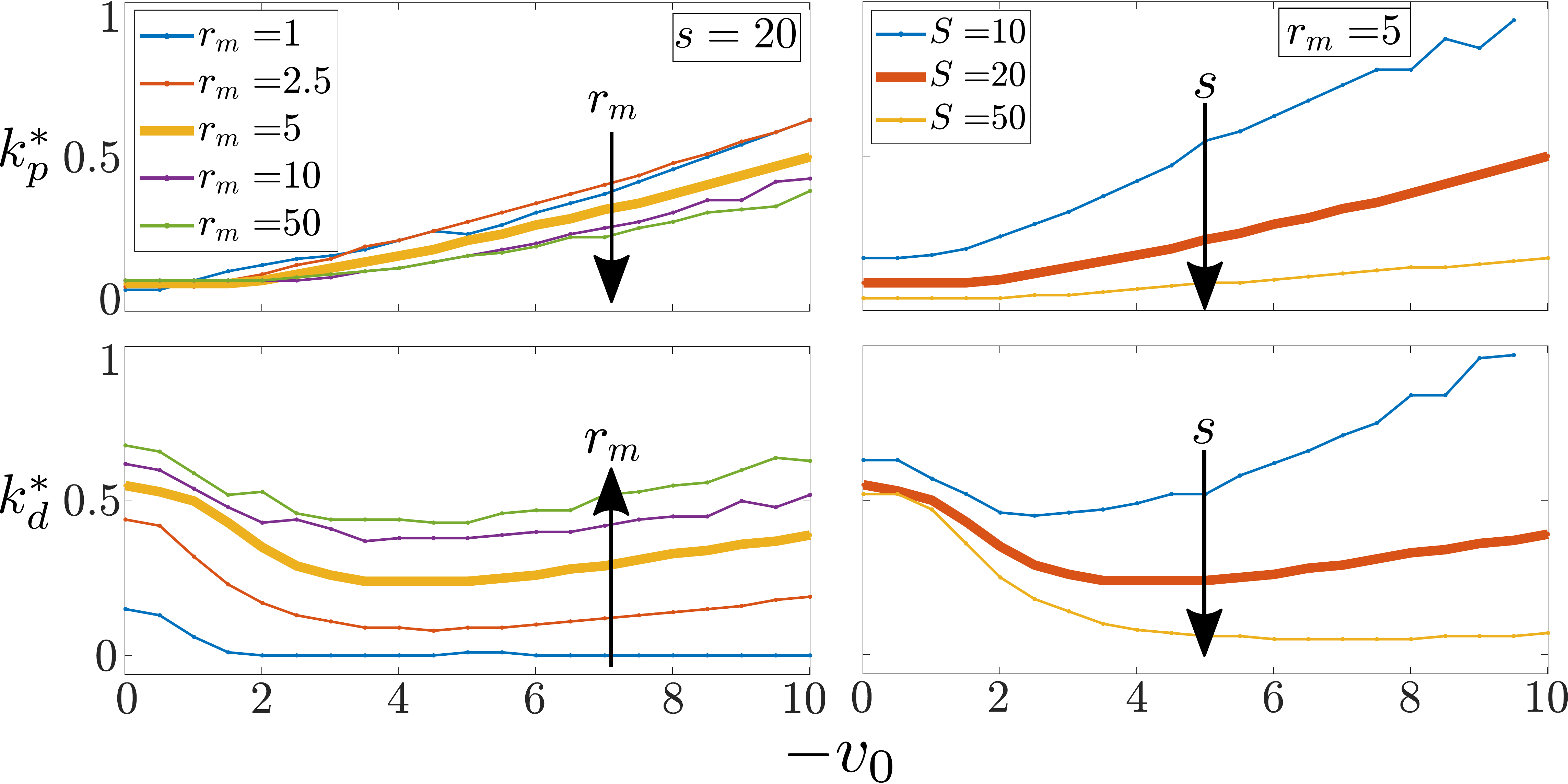}
    \caption{Dimensionless model parameters $r_m$ and $s$ affect the $k_p^*$-$v_0$ and $k_d^*$-$v_0$ curves; arrows indicate direction of increasing $r_m$ and $s$.
Increasing mass ratio $r_m$ reduces $|\partial k_p^*/\partial v_0|$ and biases the $k_d^*$-$v_0$ curve toward higher $k_d$.
Increasing stroke limit $s$ reduces $|\partial k_p^*/\partial v_0|$ and reduces $|\partial^2 k_d^*/\partial v_0^2|$ for large $v_0$.
Thick curves correspond to $r_m = 5$ and $s = 20$, the parameter values used in experiments (see Section~\ref{sec:experiments}).}
    \label{fig:opt_imp_vary_rm_S}
\end{figure}

Having analyzed the $k_p^*$-$k_d^*$-$v_0$ relationship in detail for mass ratio $r_m = 5$ and stroke limit $s = 20$, we conclude this section by studying the effect of $r_m$ and $s$ on the optimal impedance.
Figure~\ref{fig:opt_imp_vary_rm_S} shows the $k_p^*$-$v_0$ and $k_d^*$-$v_0$ curves for a range of mass ratios and stroke limits.
Optimal stiffness $k_p^*$ appears to increase monotonically with $|v_0|$, regardless of mass ratio $r_m$ and stroke limit $s$, although the magnitude of the rate of stiffness increase $|\partial k_p^*/\partial v_0|$ appears inversely proportional to both mass ratio and stroke limit.
For small $v_0$, $k_d^*$ decreases as $v_0$ increases, until $v_0$ reaches a critical velocity (somewhere between 2 and 6, depending on $s$), at which point $k_d^*$ increases with $v_0$, with its rate of increase inversely proportional to $s$.
Additionally, increasing $r_m$ appears to bias the $k_d^*$-$v_0$ curve upward.

\section{Comparison between impedance control and force control}\label{sec:bang_bang_vs_imp}

\begin{figure}[t]
    \centering
    \includegraphics[width=0.55\linewidth]{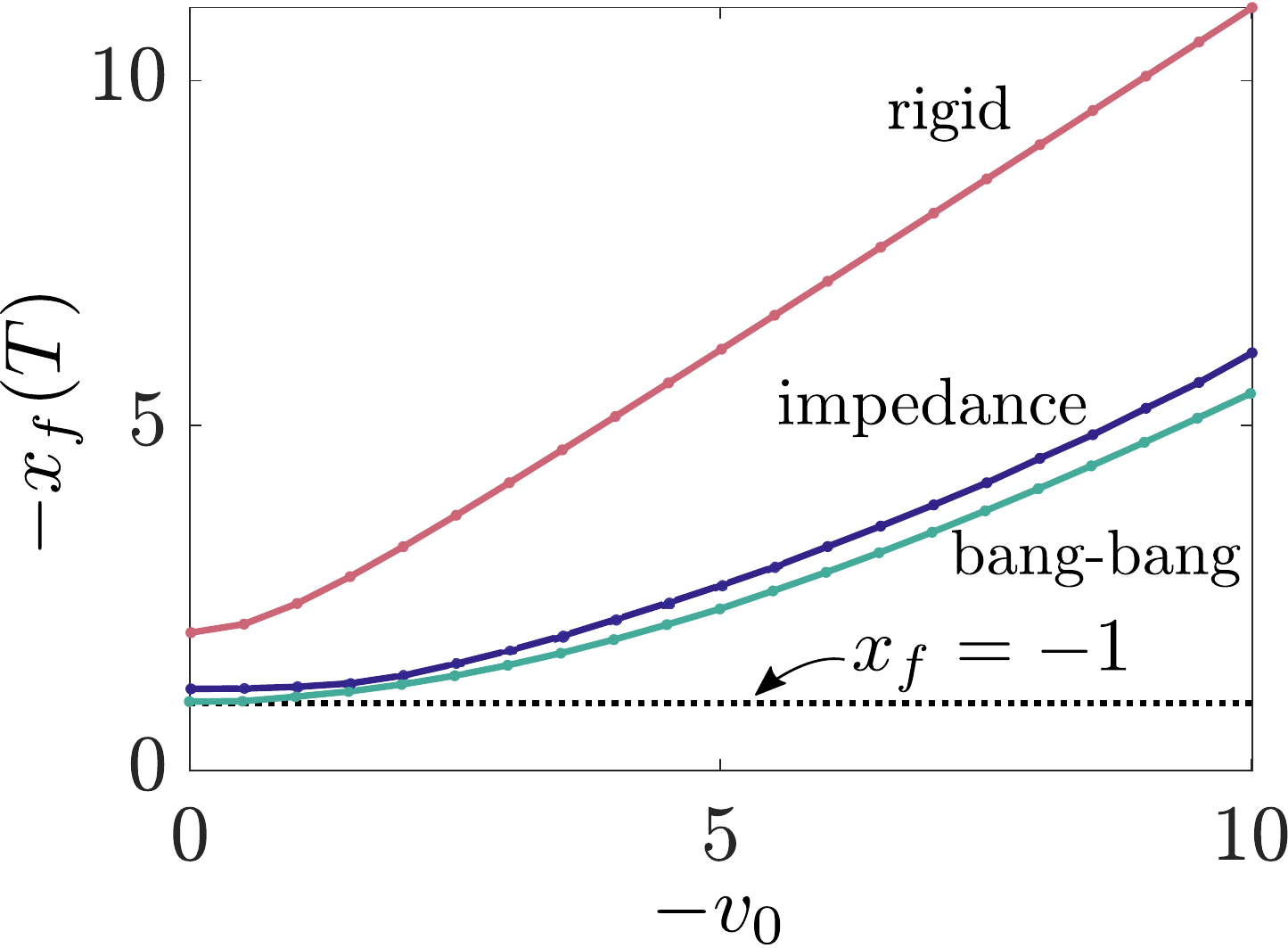}
    \caption{Nondimensionalized penetration depth $x_f$ \vs impact velocity $v_0$ for rigid, impedance-controlled, and bang-bang force-controlled leg; $r_m$ = 5, $s = 20$.
The smallest possible penetration depth is $x_f = -1$.}
    \label{fig:imp_vs_bb_vs_rigid}
\end{figure}

Depth-versus-impact velocity curves for optimal impedance control and bang-bang force control are shown in Figure~\ref{fig:imp_vs_bb_vs_rigid} along with a worst-case scenario in which the robot impacts as a rigid body.
While bang-bang force control uses the full force available ($\pm u_\textrm{max}$), optimal impedance control does not saturate $u$ until $v_0$ is sufficiently large.
For this comparison, we take $u_\textrm{max}$ to be the maximum force applied by the optimal impedance controller at the maximum simulated impact velocity ($v_0 = -10$); this results in $u_\textrm{max} = 8.2$.
Recall from Section~\ref{sec:nondimensionalization} that the minimum achievable foot penetration depth is $x_f = 1$ when $v_0 = 0$.
If the body and foot are rigidly connected, when the robot ``impacts'' with $v_0 = 0$ it will gain momentum as it sinks, penetrating to a depth of $x_f = 2$.
As $v_0$ approaches zero, foot penetration under bang-bang control approaches the best-case penetration depth $x_f = 1$; optimal impedance control results in slightly deeper foot penetration but still comes close to $x_f = 1$.
As $v_0$ grows, $\partial x_f/\partial v_0$ approaches unity more slowly for the impedance-controlled and bang-bang-controlled cases than for the rigid case.

As shown in Figure~\ref{fig:imp_vs_bb_vs_rigid}, both control policies significantly reduce penetration depth, compared to the rigid impact depth.
The open-loop force profile represents a brittle optimizer, due to the absence of feedback, whereas impedance control is more robust to model uncertainty because it has feedback.
This robustness compensates for the small increase in penetration depth compared to open-loop bang-bang force control.

\section{Experimental Validation}\label{sec:experiments}

\begin{figure*}[t]
\centering
\smallskip
\begin{subfigure}[t]{0.51\columnwidth}
\includegraphics[width=\columnwidth]{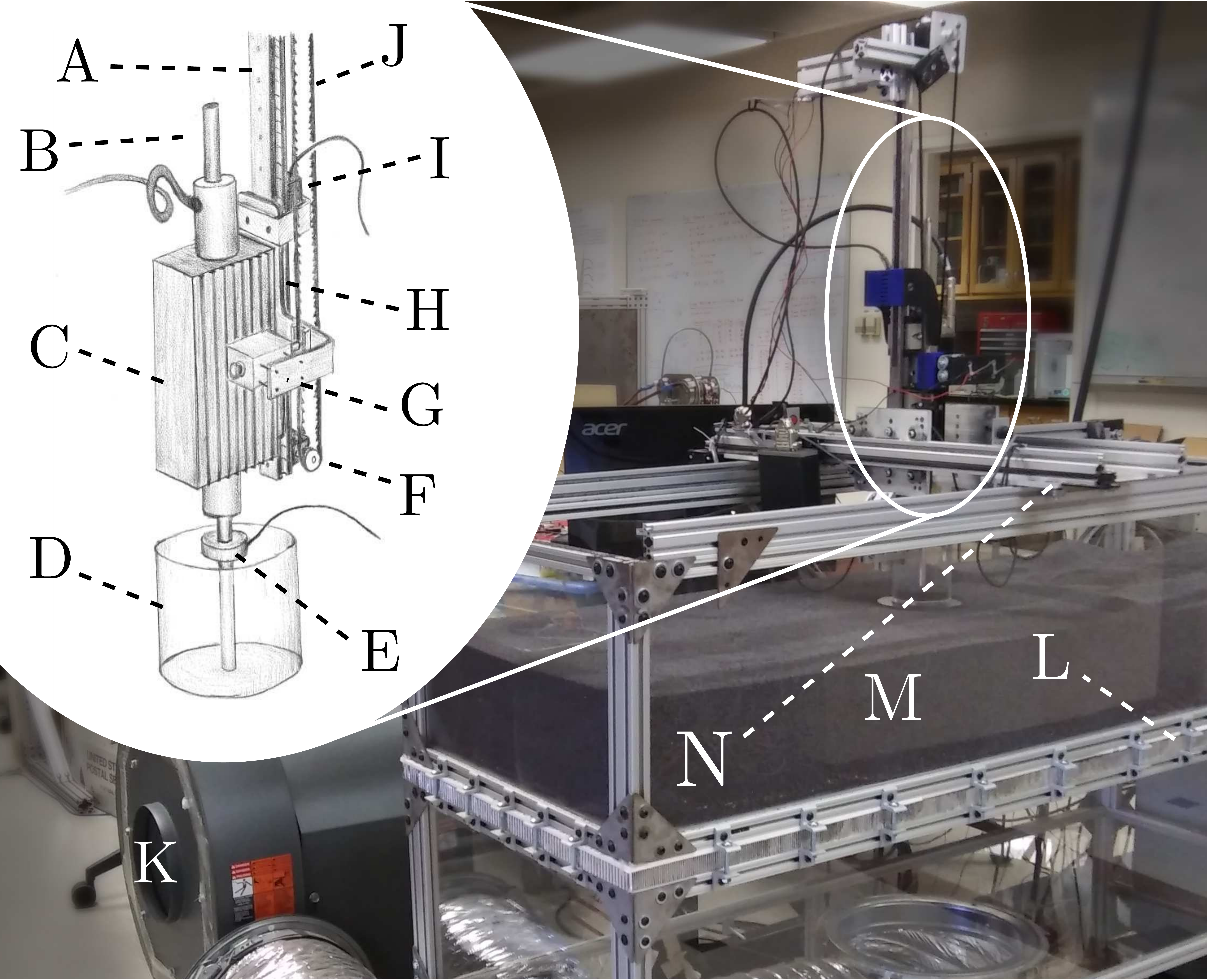}
\caption{Experimental apparatus.}%
\label{fig:apparatus_pencil_sketch}%
\end{subfigure}\hfill%
\begin{subfigure}[t]{1.48\columnwidth}
\includegraphics[width=\textwidth]{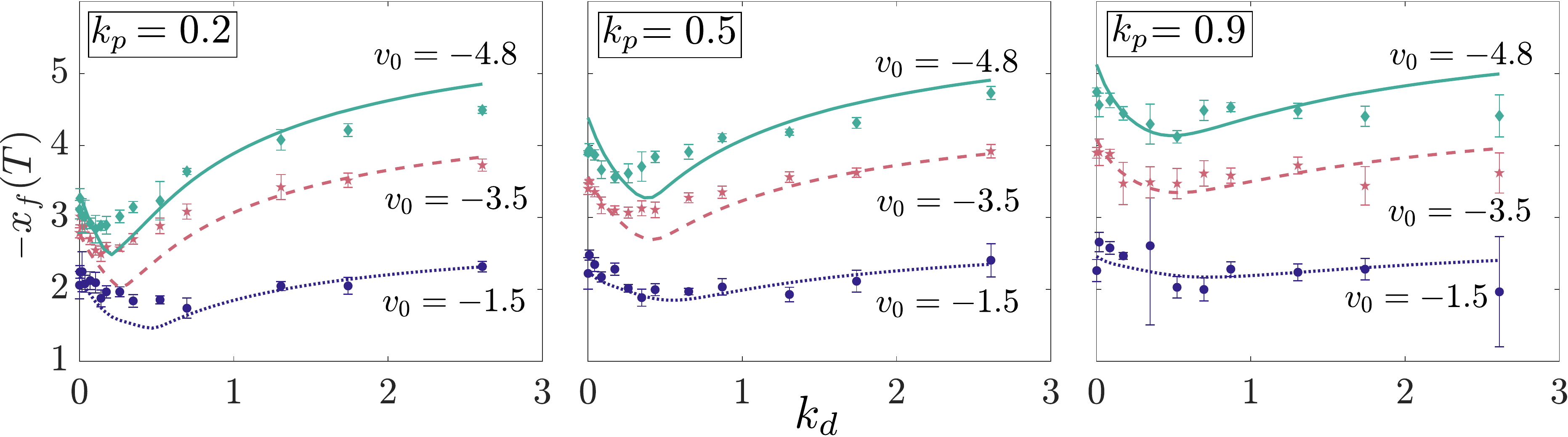}
\caption{Experimental and simulation results.}%
\label{fig:nd_exp_results}%
\end{subfigure}
\caption{Experimental apparatus and results.
\textbf{a) }\textit{1-D robot}: guiderail (A), slider (B), stator (C), hollow acrylic cylinder (D), force/torque sensor (E).
\textit{Clutch/lifting mechanism}: stepper motor (F), solenoid (G), linear carriage (H), absolute position encoder (I), and timing belt (J).
\textit{Fluidized bed trackway}: blower (K), porous membrane diffuser and honeycomb support (L), poppy seeds (M), and x-y gantry (N).
\textbf{b) } Nondimensionalized penetration depth $-x_f(T)$ \vs $k_d$ for several $k_p$ and $v_0$ ($r_m = 5$, $s = 20$).
Error bars indicate $\pm 1$ standard deviation.
}
\label{fig:exp_all}
\end{figure*}

To validate the optimal impedance-impact velocity trends observed in simulation on a physical soft substrate, we performed experiments in which a two-mass vertically-constrained impedance-controlled robot was dropped into a prepared bed of granular media.

\subsection{Experimental Setup}

The experimental apparatus, shown in Figure~\ref{fig:apparatus_pencil_sketch}, consists of three systems:

\subsubsection{Fluidized bed trackway}
This system consists of a blower driven by a variable-frequency drive, which forces air through a porous membrane diffuser and its honeycomb support, fluidizing a 20-cm deep bed of poppy seeds, chosen for their proximity in size to naturally occurring soft substrates and for their low density which makes fluidization feasible~\cite{Qian_2015},~\cite{LiFluidized2009}.
During fluidization, airflow from the blower excites the poppy seeds into a ``bubbling'' state;
when the airflow is shut off, the seeds settle into a loosly packed state, such that intrusion results in further compaction rather than dilation.
Fluidization between experiments ensures repeatable and homogeneous ground conditions.
\subsubsection{Robot}
The 1-D hopping robot consists of a body ($m_b = 2.5$ kg) and a foot ($m_f = 0.5$ kg) and is built around a LinMot PS01-23x160H-HP-R linear brushless DC motor, driven by a 32-bit microcontroller running a position-control loop at 2 kHz which is used to emulate viscoelastic forces and apply feedforward friction and cogging compensation.
An incremental encoder inside the motor measures the position of the slider relative to the stator, and an ATI Mini45 6-axis force/torque sensor mounted in series between the slider and a hollow acrylic cylinder (130 mm diameter) measures ground reaction forces.
The diameter of the acrylic cylinder results in a ground stiffness of $k_g$ = 4.4 kN/m, determined by measuring force and penetration depth during quasistatic intrusions.
\subsubsection{Clutch and lifting mechanism} 
This system consists of a solenoid and a timing belt driven by a stepper motor and is used to suspend the robot along a vertical guiderail before dropping, to control the impact velocity.
The LinMot stator is mounted on a carriage that rides along the guiderail, and an RLS LA11 absolute magnetic linear encoder measures the position of the carriage along the guiderail.

\subsection{Experimental Procedure and Results}
We performed impedance-controlled impact experiments over a range of impact velocities and impedances.
While arbitrarily large impact velocities are achievable in simulation, experimentally achievable impact velocities do not exceed 1.2 m/s (or $v_0 = -4.8$), due to the maximum height from which the robot can be lifted and then dropped.
In simulation, the effect of damping on penetration depth is most pronounced at low stiffnesses (see Figure~\ref{fig:Kp_Kd_grid}), so we first sought the lowest position feedback gain achievable at the maximum impact velocity ($v_0 = -4.8$) without exceeding the stroke limit;\footnote{If the leg is not stiff enough, the body will collapse onto to the foot upon impact, violating the stroke limit.}
this position feedback gain corresponded to $k_p = 0.2$.
We selected $k_p = 0.5$  and $k_p = 0.9$ for the second and third relative stiffnesses, respectively.
We selected a range of $k_d$ values between 0 and 2.5, closely spaced for small $k_d$ and further apart for large $k_d$ to obtain higher resolution near the optimal solutions predicted in simulation.
We performed five impacts for each combination of $k_p$, $k_d$, and $v_0$.

Nondimensionalized penetration depth data from the experiments are plotted along with simulated depth-versus-damping curves in Figure~\ref{fig:nd_exp_results}.
Comparing the three panels of Figure~\ref{fig:nd_exp_results}, observe that for small damping $k_d$, as stiffness $k_p$ increases, so does penetration depth (for each impact velocity $v_0$);
this trend is less noticeable for larger $k_d$ as the robot begins to resemble a rigid impactor.
At low $k_p$, the optimal damping $k_d^*$ decreases as impact speed $|v_0|$ increases, but $|\partial x_f/\partial k_d|$ increases as $|v_0|$ increases.

The experiments qualitatively reflect the trends observed in simulation for $r_m = 5$ and $s = 20$, although discrepancies between experimental data and simulation are noticeable for low impedances.
This disagreement may be largely attributed to nonlinearities in the apparatus (\eg friction between the motor slider and stator, friction between the carriage and guiderail, force ripple in the motor, and/or off-axis loading during impact).
While feedforward force control is used to compensate for these nonlinearities, it is imperfect, and at low impedances these errors in applied force are larger relative to the emulated viscoelastic force and the ground reaction force.

Additionally, the unidirectional-spring GRF model is indeed a first-order approximation, and higher-order effects such as inertial drag are more pronounced for large impedances and impact velocities, exemplified by the disagreement between experiments and simulation for $v_0 = -4.8$, $k_p = 0.5, 0.9$ and large $k_d$.
Nevertheless, the experimental results confirm the impact-velocity-dependence of the penetration depth-minimizing impedance on a real example of yielding terrain.

\section{Minimum cost-of-transport hopping}\label{sec:discussion}
\begin{figure}[t]
    \centering
    \begin{subfigure}[t]{0.48\linewidth}
        \centering
        \includegraphics[width=\textwidth]{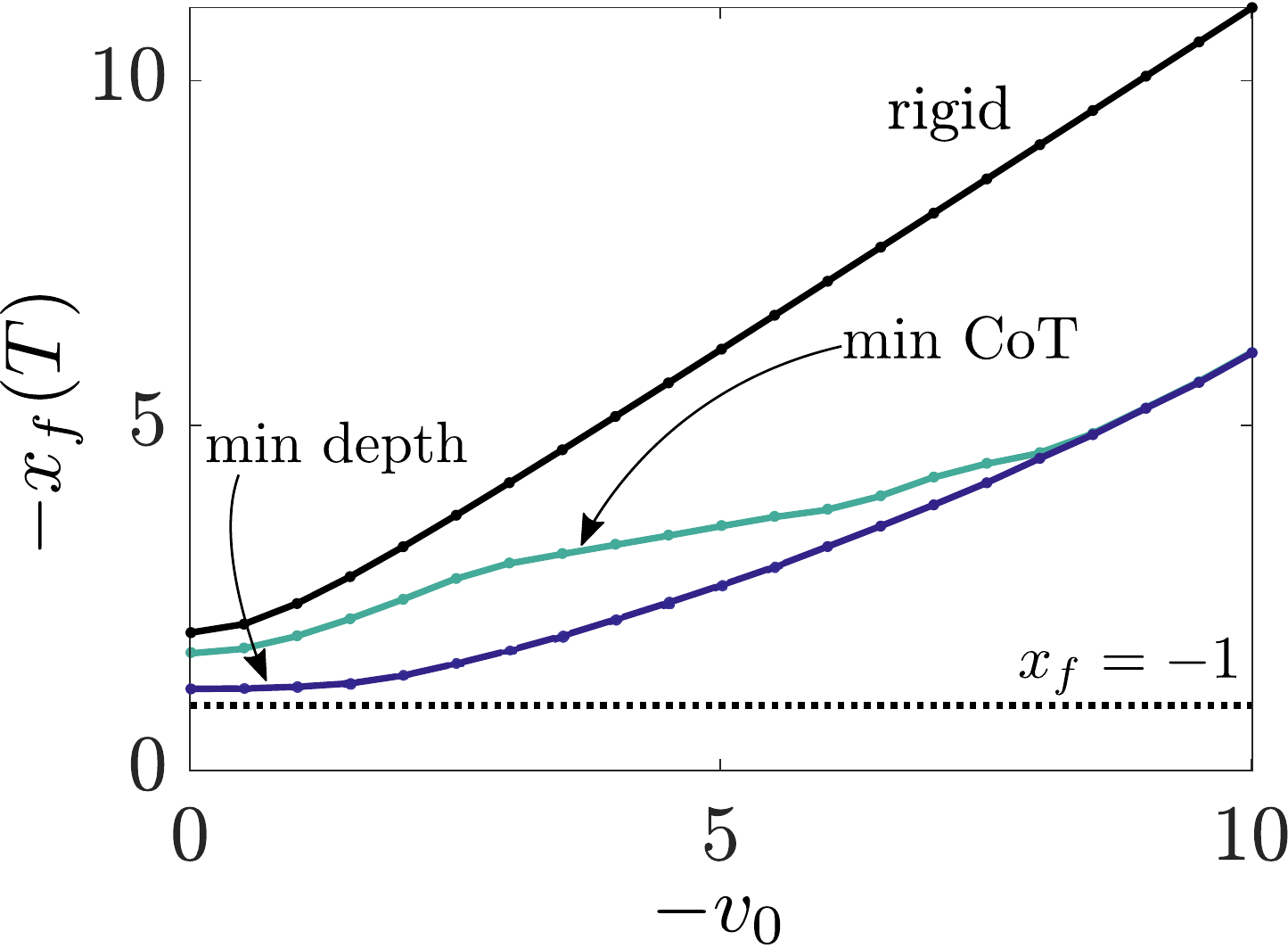}
        \caption{Depth comparison.}
        \label{fig:CoT_opt_imp_compare_depth}
    \end{subfigure}%
    ~
    \begin{subfigure}[t]{0.48\linewidth}
        \centering
        \includegraphics[width=\textwidth]{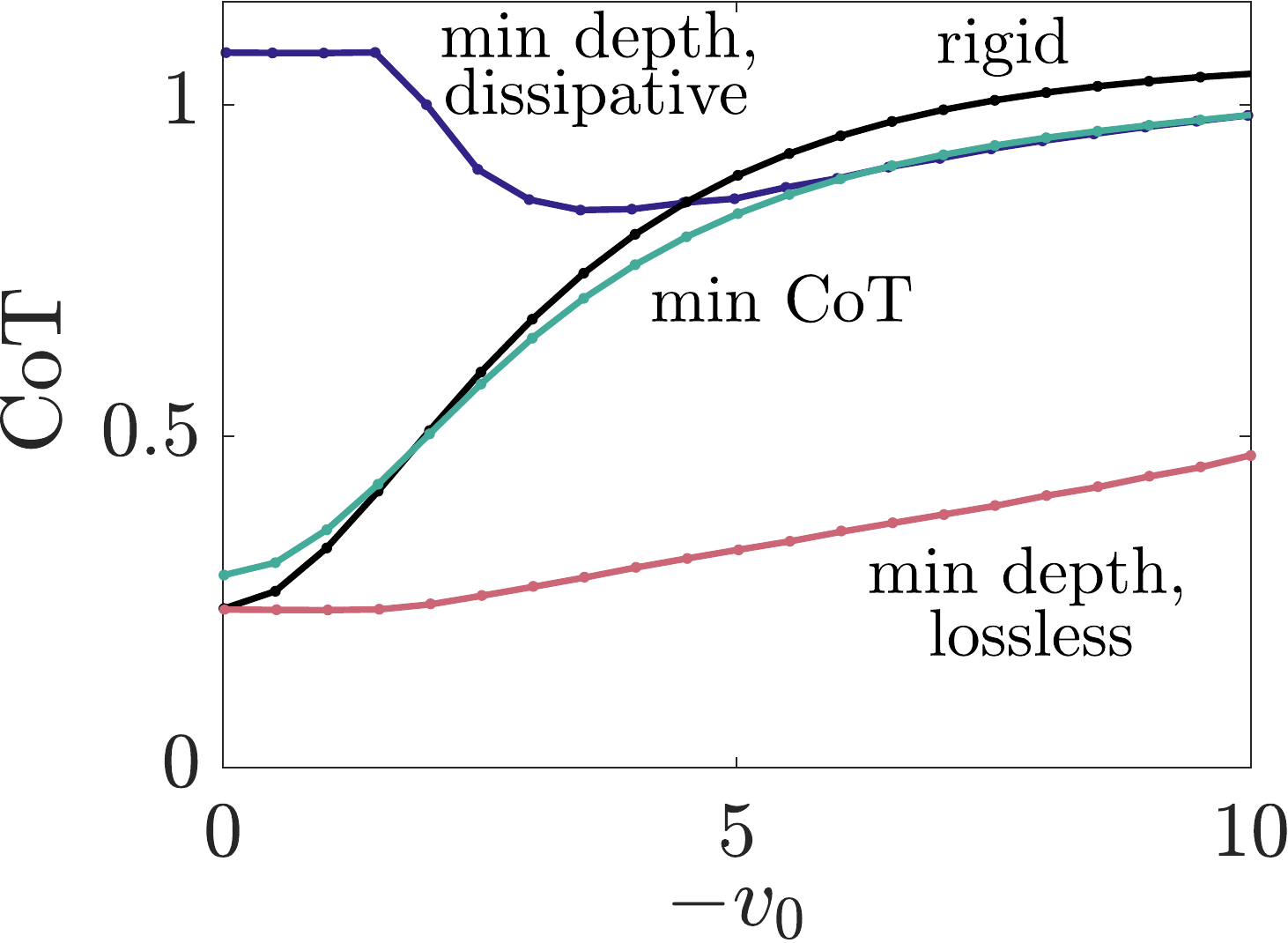}
        \caption{CoT comparison.}
        \label{fig:CoT_opt_imp_compare_CoT}
    \end{subfigure}
    \caption{
Comparison between minimum-CoT and minimum-depth impedance-controlled impacts.
\textbf{a) } For small impact velocity $v_0$, the minimum-CoT solution penetrates almost as much as the rigid impactor.
As $v_0$ grows, the minimum-CoT solution converges to the minimum-depth solution.
\textbf{b) }When viscoelastic forces are emulated through dissipative actuation $\left(\mathrm{CoT} = \left(E_\textrm{gnd} + E_\textrm{act}\right)/E(0)\right)$, the minimum-depth solution has a significantly larger CoT at small $v_0$ than the rigid impactor because actuator dissipation is larger than ground dissipation.
As $v_0$ grows, the CoT-$v_0$ curves approach the same slope, but the CoT of the minimum-depth solution is reduced significantly when impedance is rendered losslessly $\left(\mathrm{CoT} = E_\textrm{gnd}/E(0)\text{, }E_\textrm{act} = 0\right)$.
}
    \label{fig:CoT_opt_imp}
\end{figure}

We extend our work to hopping gaits by examining the relationship between minimum penetration depth and minimum cost of transport (CoT) for impedance-controlled impacts.
Our model is restricted to vertical motion, \ie there is no characteristic horizontal length scale by which to normalize energy expenditure, so instead we define cost of transport as the relative energy loss during impact:
\begin{equation}\label{eq:CoT_definition}
\text{CoT} = \frac{E(0) - E(T)}{E(0)},
\end{equation}
where $E(0)$ and $E(T)$ are the total mechanical energy of the robot-ground system at impact ($\tau = 0$) and when the foot comes to rest ($\tau = T$), respectively.\footnote{Because the flight phase of a hopping gait is governed by ballistic dynamics, $E(0)$ scales with stride length; thus, normalizing by $E(0)$ has the effect of normalizing by stride length, to within a scaling factor.
The mechanical energy lost during impact is $E(0) - E(T)$, and in a hopping gait this energy must be injected back into the robot to maintain cycle-to-cycle stability, so it is a reasonable metric of cycle-wise energy expenditure.}
The CoT can range from 0 to $\infty$, where $\mathrm{CoT} = 0$ represents a perfectly elastic impact, $\mathrm{CoT} = 1 $ represents a perfectly inelastic impact, and $\mathrm{CoT} > 1$ reflects additional energy loss due to work done by the actuator and by gravity.
Losses occur through actuation and ground dissipation: $E(0) - E(T) = E_\textrm{act} + E_\textrm{gnd}$.
From the unidirectional spring GRF model (Equation~\eqref{eq:one-way_spring_grf}), energy lost due to ground deformation is given by 
$E_{\textrm{gnd}} = \frac{1}{2}x_f^2(T)$.
Actuation loss is given by 
$E_\textrm{act} = \int_0^T u(\tau)\left(\dot{x}_b(\tau) - \dot{x}_f(\tau)\right)\mathrm{d}\tau\text{.}$
For impedance control (Equation~\eqref{eq:nondim_imp_control}), the integrand in $E_\textrm{act}$ is quadratic in $\dot{x}_b - \dot{x}_f$;
consequently, the CoT-minimizing impedance is large for low impact velocities, and the resulting state trajectory remains far from the stroke limit.
As $v_0$ grows, the CoT-minimizing impedance converges to the depth-minimizing impedance, as shown in Figure~\ref{fig:CoT_opt_imp}.

Our metric for actuation loss is conservative, and there exist more energy-efficient actuator designs for rendering viscoelastic forces.
In particular, variable-stiffness actuators~\cite{Wolf2008VSjoint} are a promising alternative to emulated compliance, in part because they can store elastic potential energy.
Regenerative braking converts mechanical energy to electrical energy by generating a voltage proportional to motor speed~\cite{ME333book}, offering an energy efficient realization of viscous damping.
As actuator efficiency improves, $E_\textrm{act}$ shrinks relative to $E_\textrm{ground}$, so minimizing CoT increasingly depends on minimizing penetration depth.

\section{Conclusions}
In this paper we examined force control and impedance control solutions to the soft landing problem, minimizing foot penetration depth into a soft substrate for a given impact velocity.
We derived a dimensionless model of a simple robot consisting of a body and foot impacting into yielding terrain approximated as a unidirectional spring.
From this model, we formulated a constrained optimal control problem, from which we obtained an open-loop bang-bang force profile that appears to minimize foot penetration depth but is brittle due to the lack of feedback.
Motivated by biology, recent actuation trends, and the need for a more robust control policy, we also examined impedance control, seeking optimal impact-velocity-dependent stiffness and damping.
Impedance control experiments, in which a vertically-constrained two-mass robot impacted into a bed of granular media, reflected the optimal impedance solutions found in simulation, suggesting that real-world legged locomotors can indeed reduce foot sinkage with the right leg stiffness and damping.
Lastly, we looked beyond the soft landing problem to energy-efficient locomotion, seeking impedances that minimized the relative energy loss during impact.
As joint actuation technology improves in efficiency, minimizing this relative energy loss will increasingly depend on minimizing foot penetration depth.
We plan several extensions of this work, including online terrain parameter estimation for adaptive locomotion control as well as planar and 3D walking and running on yielding terrain.

\section*{Acknowledgments}
The authors wish to thank Igal Alterman for his work developing the fluidized bed trackway,
Blake Strebel for his extensive work developing the hopping robot,
and Andrew Lin for his work developing the clutch and lifting mechanism.

\bibliographystyle{IEEEtranN}
{\small
\bibliography{references}}

\end{document}